\documentclass[letterpaper]{article}
\usepackage{aaai2027}
\usepackage[hyphens]{url}
\usepackage{graphicx}
\usepackage{amsmath}
\usepackage{natbib}
\usepackage{caption}
\usepackage{algorithm}
\usepackage{algorithmic}
\usepackage{booktabs}
\usepackage{makecell}
\usepackage{array}
\newcolumntype{L}[1]{>{\raggedright\arraybackslash}p{#1}}
\title{HARGO: Heterogeneity-Aware Reward-Guided Optimization for RL Post-Training of LLMs on HPC Tasks}
\author{Tiangang Li\textsuperscript{*}\thanks{\textsuperscript{*}Corresponding author.} \quad Xiangbo Tian}
\affiliations{School of Computer Science, Wuhan University, Wuhan, China\\
\{tiangangli, tianxiangbo\}@whu.edu.cn}

\begin{document}

\maketitle

\begin{abstract}
Supervised fine-tuning (SFT) can equip large language models (LLMs) with domain knowledge for high-performance computing (HPC) tasks such as data race detection and benchmark question answering. However, knowledge alone does not guarantee task-appropriate behavior: the same SFT model that correctly classifies 88.65\% of C/C++ data race samples produces verbose, imprecise answers to factual queries, with 65.9\% of MLPerf responses exceeding 40 characters. Reinforcement learning (RL) post-training addresses this gap by optimizing for task-specific rewards rather than token-level imitation. Yet HPC tasks exhibit extreme heterogeneity, with binary classification, factual QA, and semantic generation differing by 58x in answer length, spanning three distinct reward distributions, and showing widely varying SFT accuracy. This makes uniform-weight RL methods such as GRPO suboptimal. We propose HARGO, Heterogeneity-Aware Reward-Guided Optimization, which introduces per-response importance weighting via confidence-modulated advantage: computing a discrimination signal from group-level reward contrast and a confidence signal from reference model log-probabilities, then modulating the advantage before computing per-response weights, without requiring task-type labels. Across four HPC tasks and nine methods, HARGO achieves the best performance on all three primary metrics: WinRate 54.62\%, Data Race F1 91.30\%, and PLP Similarity 0.8558. Ablation confirms complementary contributions from both signals. HARGO establishes the best overall alignment quality among compared methods for heterogeneous HPC tasks.
\end{abstract}


\section{Introduction}

Supervised fine-tuning (SFT), as demonstrated by HPC-GPT \cite{ref1}, can equip large language models (LLMs) with the domain knowledge needed for high-performance computing (HPC) tasks such as data race detection and HPC benchmark question answering. By fine-tuning a LLaMA-13B model on automatically generated HPC instruction data, HPC-GPT showed that SFT injects HPC-specific knowledge into an LLM, achieving competitive accuracy on tasks ranging from binary data race classification to factual system specification queries. However, knowledge alone does not guarantee task-appropriate behavior. The same SFT model that correctly classifies 88.65\% of data race samples produces verbose, imprecise answers to factual queries: 65.9\% of its MLPerf responses exceed 40 characters, sacrificing conciseness for stylistic imitation of the training data. This reveals a fundamental gap between knowing HPC concepts and applying them correctly across heterogeneous task formats---a gap that SFT, by design, cannot close.

SFT's teacher-forcing objective treats all reference answers as equally informative, unable to distinguish concise accurate responses from verbose partially correct ones. It cannot exploit comparative signals across multiple candidate responses to the same prompt, nor target specific error patterns such as false positives in data race detection.

Reinforcement learning (RL) post-training \cite{ref22,ref23,ref26} addresses these limitations by scoring generated responses through a reward function, computing per-response advantages that reward better answers relative to their peers. A KL-divergence penalty with respect to a frozen reference model balances knowledge retention with behavioral optimization. In the HPC domain, reward functions can be precisely defined (e.g., binary accuracy for race detection, exact string matching for factual QA), providing high-quality training signals for RL.

However, applying RL post-training to HPC tasks reveals \textbf{task heterogeneity}---four HPC tasks differ by 58$\times$ in answer length, span three reward distribution types, and show SFT accuracy from 51\% to 100\%. Uniform-weight RL methods such as GRPO cannot adapt, as the same training loop must process batch types with vastly different learning signals. This points to a broader issue: current RL scaling implicitly assumes all rollouts contribute equally, so better performance requires more compute. HARGO instead improves how effectively each unit of compute is used, adaptively reweighting gradient contributions to extract more effective learning from the same budget.

\begin{figure}[t]
\centering
\includegraphics[width=0.9\columnwidth]{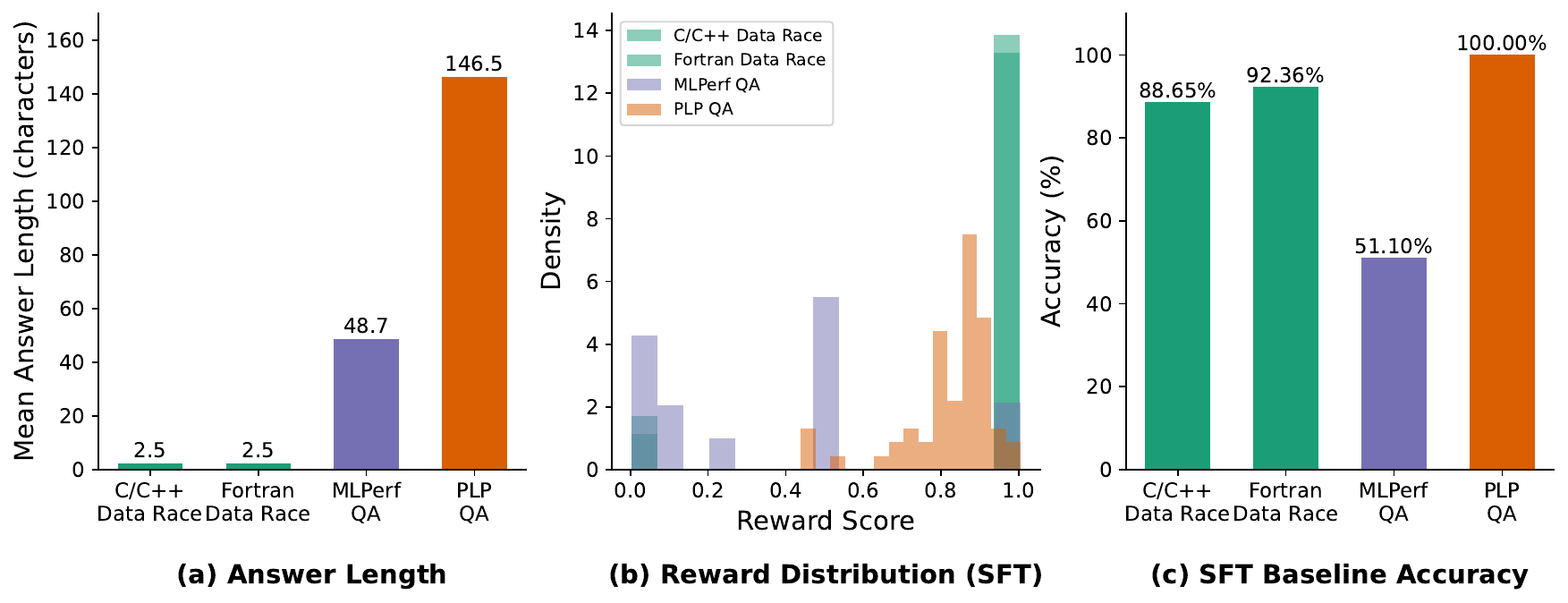}
\caption{HPC task heterogeneity across three dimensions---answer length distribution, reward distribution, and SFT baseline accuracy.}
\label{fig1}
\end{figure}

We propose \textbf{HARGO} (Heterogeneity-Aware Reward-Guided Optimization), which introduces per-response importance weighting via two complementary signals: \textbf{discrimination ($d$)}, capturing group-level reward contrast, and \textbf{confidence ($c$)}, from reference model log-probabilities. HARGO computes a modulated advantage $A_{\text{mod},i} = A_i \times (1 + \alpha \cdot c_i)$ ($\alpha = 0.3$), derives $d_i$ from $|A_{\text{mod},i}|$, and weights responses by $w_i \propto d_i$. This ensures $c$ modulates but never reverses the learning signal, with all signals computed internally without task-type labels.

To evaluate HARGO, we compare nine methods (seven RL variants plus the SFT initialization and HPC-GPT baseline) on four HPC tasks spanning binary classification, factual QA, and semantic generation, all starting from the same Qwen2.5-0.5B-Instruct checkpoint. We define three primary metrics---WinRate, Data Race F1, and PLP Similarity---with EM and AvgScore as auxiliary.

Our main results are as follows. HARGO achieves the best performance on all three primary metrics: WinRate 54.62\% (+4.79 over HPC-GPT), Data Race F1 91.30\% (+2.48), and PLP Similarity 0.8558 (+0.050). On per-task accuracy, HARGO leads on race\_fortran (94.90\%) and plp (100.00\%), and ranks third on mlperf (56.04\%, behind GRPO at 59.34\% and KTO at 56.59\%). On the auxiliary metrics EM and AvgScore, KTO achieves the best performance (EM 27.47\%, AvgScore 0.4537), with HARGO achieving competitive but not leading results (EM 17.58\%, AvgScore 0.4000). Ablation experiments confirm that both the $d$ and $c$ signals independently contribute to HARGO's performance, with the full advantage-modulation combination outperforming either signal alone.

The contributions of this paper are threefold:
\begin{enumerate}
\item We identify and quantify \textbf{HPC task heterogeneity} as a critical challenge for RL post-training---four HPC tasks differ across three measurable dimensions (answer length by 58$\times$, reward distribution type, and SFT accuracy from 51\% to 100\%)---and demonstrate that uniform-weight RL methods cannot adapt to this heterogeneity.
\item We propose \textbf{HARGO}, a per-response importance weighting method that uses confidence-modulated advantage ($A_{\text{mod}} = A \times (1 + \alpha \cdot c)$) to amplify informative responses, requires no task-type labels, and achieves consistent improvements across all three primary metrics over standard GRPO.
\item We conduct a \textbf{systematic nine-method comparison} of RL post-training approaches on HPC tasks---to our knowledge a comprehensive evaluation of its kind---establishing HARGO as the best-performing method with leading results on all three primary metrics.
\end{enumerate}

\section{Related Work}

\subsection{Large Language Models for High-Performance Computing}

While general-purpose LLMs---including GPT-4 \cite{ref29}, LLaMA \cite{ref30}, and Qwen \cite{ref31}---demonstrate strong code generation, they underperform on HPC-specific tasks. The dominant strategy is \textbf{domain-specific knowledge injection} through supervised fine-tuning. HPC-GPT \cite{ref1} pioneered this by fine-tuning LLaMA-13B on HPC instruction data; LM4HPC \cite{ref2} proposed a complementary framework. Subsequent work spans code generation \cite{ref3,ref4,ref18,ref19,ref20,ref21}, assistance tools \cite{ref5,ref6,ref7,ref8}, evaluation \cite{ref9,ref10,ref11}, and RAG/prompt-based detection \cite{ref12,ref13,ref14}, with surveys \cite{ref15,ref16,ref17} covering this intersection. The common focus is \textbf{knowledge acquisition}. None have explored whether RL post-training can close the gap between knowing HPC concepts and applying them correctly across heterogeneous tasks. Our work investigates this direction using HPC-GPT data on Qwen2.5-0.5B for controlled comparison.

\subsection{Reinforcement Learning Post-Training for Language Models}

RL post-training aligns LLM behavior after SFT using PPO with a KL penalty \cite{ref22}. Preference-based methods bypass reward models: DPO \cite{ref23} uses chosen-rejected pairs, KTO \cite{ref24} learns from unpaired preferences, SimPO \cite{ref25} eliminates the reference model. These depend on preference quality; DPO and SimPO underperform on HPC tasks (Section~\ref{sec:main_results}). Online methods score responses during training: GRPO \cite{ref26} computes group-relative advantages without a critic, proving effective for reasoning \cite{ref27}; DrGRPO \cite{ref28} removes normalization. PPO value functions are unreliable at the 0.5B scale. A common property is equal gradient contribution---GRPO assigns $1/G$ per response---which fails under HPC's heterogeneity (Section~\ref{sec:hpc_heterogeneity}). HARGO extends online RL with \textbf{per-response importance weighting}: using discrimination ($d$) from reward contrast and confidence ($c$) from reference model log-probabilities, without task-type labels.

\section{HPC Task Heterogeneity}
\label{sec:hpc_heterogeneity}

The four HPC tasks employed in this study---derived from the HPC-GPT instruction dataset \cite{ref1} and comprising data race detection benchmarks from DataRaceBench \cite{ref32}---span fundamentally different problem types, answer formats, and reward structures. Table~\ref{tab:task_overview} summarizes their key characteristics.

\begin{table}[t]
\centering

\begin{tabular}{@{}L{1.3cm} L{4.0cm} L{2.5cm}@{}}
\toprule
Task & Description & Answer type \\
\midrule
race\_c & C/C++ OpenMP data race detection & Binary (yes/no) \\
race\_fortran & Fortran OpenMP data race detection & Binary (yes/no) \\
mlperf & HPC benchmark factual QA & Factual (numbers/names/short text) \\
plp & Programming language processing descriptive QA & Descriptive (long text) \\
\bottomrule
\end{tabular}
\caption{HPC task overview. The HPC-GPT open-source instruction dataset \cite{ref1} covers four heterogeneous tasks. Data is split 9:1 by task; evaluation samples are held out. Dataset sizes and sources are detailed in Section~\ref{sec:exp_setup}.}
\label{tab:task_overview}
\end{table}

These tasks differ along three dimensions as illustrated in Figure~\ref{fig1}---answer length (up to 58$\times$), reward distribution type, and SFT baseline accuracy---creating a structural challenge for group-based RL methods.

In GRPO \cite{ref26}, uniform weighting $(1/G)$ treats all responses identically, yet the same loop processes both race batches (mostly correct, weak gradients) and mlperf batches (informative advantages). A weighting scheme treating all responses identically cannot exploit this diversity. This motivates HARGO: per-response weighting via two complementary signals that automatically adapt gradient contributions.

\section{HARGO: Heterogeneity-Aware Reward-Guided Optimization}

As established in Section~\ref{sec:hpc_heterogeneity}, the four HPC tasks exhibit extreme heterogeneity across answer length (58$\times$), reward distribution type, and SFT baseline accuracy (51\%--100\%). GRPO's uniform weighting strategy treats every response identically regardless of these differences. HARGO addresses this by introducing per-response importance weighting via two complementary, internally computed signals: \textbf{discrimination} ($d$), which captures group-level reward contrast, and \textbf{confidence} ($c$), derived from reference model log-probabilities. Together these signals automatically adapt each response's gradient contribution based on its learning value, without requiring task-type labels or prior knowledge.

\subsection{Preliminaries: GRPO}

A central observation motivating HARGO is that not all rollouts within a training group are equally informative. Under GRPO's uniform weighting, a rollout that arrives at the correct answer through careful reasoning and one that guesses correctly by chance receive identical gradient contributions, as long as their reward scores match. This is a compute-efficiency assumption: the only way to extract more learning signal is to sample more rollouts. HARGO challenges this assumption by asking whether we can improve performance at fixed group size by identifying which rollouts genuinely carry the most information for policy improvement. The method's two signals---discrimination and confidence---are designed to answer this question, reallocating gradient resources toward rollouts with the highest learning value.

Group Relative Policy Optimization (GRPO) \cite{ref26} is an online RL post-training method that eliminates the need for a learned value function by computing advantages within groups of responses. For a given prompt $q$, GRPO samples $G$ responses $\{y_1, \dots, y_G\}$ from the current policy $\pi_{\theta}$ and scores each response with a reward function $R$ to obtain rewards $\{r_1, \dots, r_G\}$. The group-relative advantage for response $y_i$ is:

\begin{equation}
A_i = \frac{r_i - \bar{R}}{\sigma_R + \varepsilon}, \quad
\bar{R} = \frac{1}{G} \sum r_i, \quad
\sigma_R = \text{std}(r_1, \dots, r_G)
\end{equation}

where $\varepsilon$ is a small constant preventing division by zero. The per-response loss combines a clipped policy gradient surrogate with a KL-divergence penalty against a frozen reference model $\pi_{\text{ref}}$ (the SFT initialization):

\begin{equation}
\begin{aligned}
L_i = \text{mean}_t\Bigl(&-\min\!\bigl(\rho_t A_i,\; \text{clip}(\rho_t, 1-\epsilon, 1+\epsilon)\, A_i\bigr) \\
&+ \beta \cdot \text{KL}(\pi_{\theta} \,\|\, \pi_{\text{ref}})\Bigr)
\end{aligned}
\end{equation}

where $\rho_t = \pi_{\theta}(y_{i,t} \mid y_{i,<t}, q)\, /\, \pi_{\theta_{\text{old}}}(y_{i,t} \mid y_{i,<t}, q)$ is the per-token importance ratio. The final GRPO loss averages equally over the $G$ responses:

\begin{equation}
L_{\text{GRPO}} = \frac{1}{G} \sum_i L_i
\end{equation}

The equal-weight average is the key limitation: every response contributes $1/G$ to the gradient regardless of whether it is a critical learning point (clearly wrong, providing a strong corrective signal) or a trivial case (just one of several correct answers in an all-correct group). As Section~\ref{sec:hpc_heterogeneity} demonstrated, this uniform strategy wastes capacity on uninformative batches while failing to amplify genuinely informative responses.

\subsection{HARGO: Per-Response Importance Weighting}

HARGO replaces GRPO's uniform average with a \textbf{weighted sum}:

\begin{equation}
L_{\text{HARGO}} = \sum_i w_i \cdot L_i, \quad \text{where} \quad \sum_i w_i = 1
\end{equation}

The weights $w_i$ are computed from two complementary signals that together assess the learning value of each response $y_i$.

\subsubsection{Discrimination Signal ($d$) via Advantage Modulation}

HARGO extends GRPO by modulating the group-relative advantage with the confidence signal before computing per-response weights. Given the standard GRPO advantage $A_i = (r_i - \bar{R}) / (\sigma_R + \varepsilon)$, we define the modulated advantage:

\begin{equation}
A_{\text{mod},i} = A_i \times (1 + \alpha \times c_i)
\end{equation}

where $c_i \in (0, 1)$ is the reference model confidence (Section~\ref{sec:confidence}) and $\alpha = 0.3$ controls the modulation strength. The modulated advantage preserves the sign of $A_i$ (direction of the gradient update) while scaling its magnitude by a factor between $1.0$ (when $c_i \approx 0$) and $1.3$ (when $c_i \approx 1.0$, $\alpha = 0.3$). This design ensures that the confidence signal modulates---but does not reverse---the learning signal: a response with high confidence receives up to 30\% more gradient amplification than the same response under standard GRPO, while a low-confidence response is effectively penalized.

The per-response discrimination signal $d_i$ is then defined via normalized modulation magnitude:

\begin{equation}
d_i = \frac{|A_{\text{mod},i}|}{\max_j |A_{\text{mod},j}| + \varepsilon}
\end{equation}

where $\varepsilon$ prevents division by zero. In all-correct groups ($\sigma_R \approx 0$, $A_i \approx 0$ for all $i$), $A_{\text{mod},i} \approx 0$ and $d_i \approx 0$ for all responses, triggering the equal-weight fallback described in Section~\ref{sec:per_response_weights}.

This formulation differs critically from additive combination ($w_i = d_i + \alpha \cdot c_i$): instead of treating $c$ as a co-equal signal added to the weight, advantage modulation embeds $c$ within the advantage computation. An alternative multiplicative scheme $w_i = d_i \times c_i$ was also considered: under multiplication, when $c_i$ and $d_i$ are anti-correlated (a high-$d_i$ response with low $c_i$), the product significantly suppresses the primary $d$ signal, distorting weight ordering. Additive combination $w_i = d_i + \alpha \cdot c_i$ avoids this distortion but introduces a positive feedback loop---the gradient always exceeds either $d_i$ alone or $c_i$ alone, amplifying policy updates and potentially destabilizing training. Advantage modulation resolves both issues: the gradient is bounded to at most $(1+\alpha)$ times the base advantage (30\% above GRPO), and $c$ modulates magnitude without altering $d_i$'s rank order.

\subsubsection{Confidence Signal ($c$)}
\label{sec:confidence}

The discrimination signal alone cannot distinguish between two subtly different scenarios: a response that is incorrect \textit{and the model already knows it} (low reference log-probability) versus a response that is incorrect \textit{contrary to the model's learned knowledge} (high reference log-probability). The former may be random noise; the latter is a genuine mistake that merits correction.

To capture this distinction, HARGO introduces a \textbf{confidence signal} $c$ derived from the reference model's per-token log-probability. For each response $y_i$, we first compute the per-token mean log-probability under the frozen reference model:

\begin{equation}
\text{ref\_logp}_i = \frac{1}{|y_i|} \sum_{t=1}^{|y_i|} \log \pi_{\text{ref}}(y_{i,t} \mid y_{i,<t}, q)
\end{equation}

To make this signal comparable across tasks with different log-probability scales, we maintain an exponential moving average (EMA) of $\text{ref\_logp}_i$ across batches:

\begin{equation}
\begin{aligned}
\text{ref\_logp}_{\text{global}} \leftarrow\;& \rho \cdot \text{ref\_logp}_{\text{global}} \\
&+ (1 - \rho) \cdot \text{mean}_{\text{batch}}(\{\text{ref\_logp}_i\})
\end{aligned}
\end{equation}

where $\rho = 0.9$. The confidence signal is then defined as the sigmoid-normalized deviation from this global baseline:

\begin{equation}
\begin{aligned}
c_i &= \sigma(\text{ref\_logp}_i - \text{ref\_logp}_{\text{global}}) \\
    &= \frac{1}{1 + \exp(-(\text{ref\_logp}_i - \text{ref\_logp}_{\text{global}}))}
\end{aligned}
\end{equation}

This formulation has three desirable properties. First, $c_i \in (0, 1)$, providing a stable reference signal independent of the reward distribution. Second, responses where the reference model is more confident than average ($\text{ref\_logp}_i > \text{ref\_logp}_{\text{global}}$) receive $c_i > 0.5$, modestly increasing their weight---these are likely consistent with the SFT-learned HPC knowledge. Third, the EMA baseline adapts dynamically during training: as the policy improves, the global log-probability baseline shifts, providing a moving reference point for what counts as ``confident.'' This avoids the brittleness of a fixed absolute threshold.

\subsection{Per-Response Weights and Design Decisions}
\label{sec:per_response_weights}

The per-response weight is directly proportional to $d_i$, with equal-weight fallback for zero-advantage groups:

\begin{equation}
w_i = \frac{d_i}{\sum_j d_j} \quad \text{(normalized weight)}
\end{equation}
\begin{equation}
\text{if } \sum_j d_j < \varepsilon_w \text{ then } w_i = \frac{1}{G} \text{ for all } i \quad \text{(equal fallback)}
\end{equation}

This scheme has three key properties. \textbf{First}, advantage modulation bounds the gradient to at most $1+\alpha$ times GRPO, avoiding additive combination's feedback loop. \textbf{Second}, $\alpha = 0.3$ was selected via sweep over $\{0.1, 0.3, 1.0\}$: $\alpha = 0.1$ is too weak, $\alpha = 1.0$ allows $c$ to dominate, and $\alpha = 0.3$ balances these extremes. \textbf{Third}, all-correct groups naturally trigger the equal-weight fallback, maintaining KL regularization.

\subsection{Training Procedure}

Algorithm~1 (see Supplementary Document) summarizes the complete training procedure, including confidence-modulated advantage computation and per-response adaptive weighting.

\subsection{Comparison with Related RL Methods}

HARGO occupies a specific point in the design space of RL post-training methods. Table~\ref{tab:method_comparison} contrasts HARGO with the methods evaluated in our experiments.

\begin{table*}[t]
\centering
\small
\setlength{\tabcolsep}{1.8mm}
\begin{tabular}{@{}p{3.2cm}cccccc@{}}
\toprule
\makecell{Dimension} & \makecell{DPO\\(Rafailov et al.)} & \makecell{KTO\\(Ethayarajh et al.)} & \makecell{SimPO\\(Meng et al.)} & \makecell{GRPO\\(Shao et al.)} & \makecell{DrGRPO\\(Liu et al.)} & HARGO \\
\midrule
Training paradigm & Offline & Offline & Offline & Online & Online & Online \\
Value function & No & No & No & No & No & No \\
Preference data & Yes (pairs) & Yes (unpaired) & Yes (pairs) & No & No & No \\
Reference model & Yes & Yes & No & Yes & Yes & Yes \\
Weighting strategy & Uniform & Uniform & Uniform & Uniform $(1/G)$ & Uniform $(1/G)$ & \textbf{Per-response adaptive} \\
Heterogeneity adaptation & No & No & No & No & No & \textbf{Yes ($A_{\text{mod}}$)} \\
\bottomrule
\end{tabular}
\caption{Comparison of representative RL post-training methods. We distinguish all nine methods evaluated in our experiments---PPO, DPO, KTO, SimPO, GRPO, DrGRPO, and HARGO---by their training paradigm, weighting strategy, and heterogeneity adaptation. SFT and HPC-GPT serve as non-RL baselines.}
\label{tab:method_comparison}
\end{table*}

The key architectural distinction is the weighting strategy. GRPO, PPO, DPO, and KTO all assign equal per-sample weight within a batch or group. HARGO introduces per-response importance weighting via confidence-modulated advantage ($A_{\text{mod}} = A \times (1 + \alpha \cdot c)$), derived entirely from quantities already present in GRPO-style training---requiring no additional models, task-type labels, or annotation. The $d$ signal is computed from the modulated advantage; the $c$ signal requires only the reference model's log-probabilities, which are already computed for the KL penalty term. The additional computational overhead is marginal: the EMA update (one scalar per batch) and the sigmoid computation (one per response).

\section{Experiments}

We evaluate HARGO against eight baseline methods across four HPC tasks. All experiments use a single consistent setup: every method starts from the same Qwen2.5-0.5B-Instruct \cite{ref31} base model fine-tuned on the same HPC instruction dataset, and all RL methods share an identical reward function. The only variable across methods is the training objective.

\subsection{Experimental Setup}
\label{sec:exp_setup}

\textbf{Data.} We use the HPC-GPT open-source instruction dataset \cite{ref1} (5,273 train / 584 eval, 9:1 split) across four tasks: race\_c, race\_fortran, mlperf, and plp. Race data originate from DataRaceBench \cite{ref32}; MLPerf data from the MLPerf benchmark \cite{ref34}. The dataset is at \url{https://huggingface.co/datasets/HPC-GPT/HPC}.

\textbf{Training protocol.} Stage 1 (SFT): Qwen2.5-0.5B-Instruct fine-tuned on 5,273 samples, 2 epochs (lr $2 \times 10^{-5}$, batch size 4). Stage 2 (RL): each method trained from the same SFT checkpoint. Table~\ref{tab:training_config} lists HARGO's configuration; baseline configurations are in the Supplementary Document. All experiments run on a single RTX 3080 GPU (16 GB VRAM). Random seeds are fixed. The model checkpoint post-trained with HARGO is available at \url{https://huggingface.co/swaggy/Qwen2.5-0.5B-HARGO-HPC}.

\begin{table}[t]
\centering

\begin{tabular}{@{}l c@{}}
\toprule
Hyperparameter & Value \\
\midrule
Group size $G$ & 4 \\
KL coefficient $\beta$ & 0.02 \\
Confidence coefficient $\alpha$ & 0.3 \\
EMA decay $\rho$ & 0.9 \\
Learning rate & $1 \times 10^{-5}$ \\
Epochs & 3 \\
Batch size (prompts) & 2 \\
Generation temperature $T$ & 0.6 \\
Max new tokens & 64 \\
Optimizer & AdamW \\
Gradient clip & 1.0 \\
\bottomrule
\end{tabular}
\caption{HARGO training configuration. All baseline methods are trained from the same SFT checkpoint. Their complete configurations are listed in the Supplementary Document.}
\label{tab:training_config}
\end{table}

\textbf{Reward function.} All RL methods share a single task-adaptive reward function $R(q, y)$ that evaluates generated responses without learned components. For race tasks, $R$ outputs 1.0 for exact yes/no match and 0.0 otherwise. For mlperf, $R$ assigns 1.0 for exact or numeric match, 0.5 for partial multi-value match, 0.2 for keyword overlap exceeding 70\%, 0.1 for keyword overlap exceeding 40\%, and 0.0 otherwise. EM evaluation additionally uses regex \texttt{\textbackslash b} word-boundary matching on mlperf answers to ensure token-level precision. For plp, $R$ computes cosine similarity between the generated response and the reference answer using the all-MiniLM-L6-v2 sentence-transformer model \cite{ref33}. A small format bonus (+0.05) rewards non-empty, non-repetitive responses, and a length penalty ($-0.001$ per character beyond 200) discourages verbosity. This reward function is an external, fixed component---all RL methods, including HARGO, use it identically; HARGO's contribution is in the loss weighting, not in reward design.

\textbf{Preference data.} DPO, KTO, and SimPO require preference data, which we generate automatically: for each training prompt, the SFT model produces two responses, the reward function scores both, and the higher-scoring response is labeled as chosen, the lower as rejected. No human annotation is involved.

\textbf{Evaluation metrics.} We define three primary metrics covering all four HPC tasks. \textbf{WinRate} measures the fraction of evaluation samples where an RL method's response is preferred over the SFT baseline (computed as $\text{win} + 0.5 \times \text{tie}$, expressed as a percentage), providing a global measure of alignment improvement. \textbf{Data Race F1} is the macro-averaged F1 score across race\_c and race\_fortran, computed via word-boundary extraction of yes/no judgments from model outputs---independent of the reward function. \textbf{PLP Similarity} is the mean cosine similarity (all-MiniLM-L6-v2) between generated and reference answers on plp samples, providing a quasi-independent measure of semantic fidelity. We additionally report two auxiliary metrics: \textbf{Exact Match (EM)} on mlperf, which uses lenient word-boundary matching, and \textbf{MLPerf AvgScore}, the mean reward assigned by the mlperf reward function. EM and AvgScore are not used for method ranking, as EM measures verbatim reproduction rather than alignment quality, and AvgScore is fully determined by the reward function used during training.

\subsection{Main Results}
\label{sec:main_results}

Table~\ref{tab:primary_results} presents the primary results. Each RL method is compared against the SFT baseline (the starting point for all methods) and the HPC-GPT baseline (SFT-only, included to isolate the effect of RL post-training from the effect of domain-specific fine-tuning).

\begin{table}[t]
\centering

\setlength{\tabcolsep}{1.2mm}
\begin{tabular}{@{}l c c c c c@{}}
\toprule
Method & WR$\star$ & F1$\star$ & PLP$\star$ & EM(a) & AS(a) \\
\midrule
SFT & --- & 89.59 & 0.8242 & 14.29 & 0.3587 \\
HPC-GPT & 49.83 & 88.82 & 0.8054 & 19.23 & 0.4027 \\
PPO & 50.17 & 90.22 & 0.8221 & 18.68 & 0.3779 \\
DPO & 40.24 & 77.61 & 0.7993 & 7.14 & 0.2424 \\
GRPO & 53.17 & 90.73 & 0.8351 & 18.13 & 0.4193 \\
DrGRPO & 51.63 & 90.79 & 0.8388 & 15.93 & 0.3806 \\
SimPO & 44.26 & 90.03 & 0.7816 & 6.04 & 0.1847 \\
KTO & 53.08 & 90.16 & 0.8449 & \textbf{27.47} & \textbf{0.4537} \\
\textbf{HARGO} & \textbf{54.62} & \textbf{91.30} & \textbf{0.8558} & 17.58 & 0.4000 \\
\bottomrule
\end{tabular}
\caption{Primary results. $\star$ = primary metric. Bold = best. The SFT model serves as the WinRate baseline (not directly comparable). HPC-GPT is an SFT-only model trained on the same data. All RL methods start from the identical SFT checkpoint. EM and AvgScore are auxiliary metrics.}
\label{tab:primary_results}
\end{table}

\begin{table}[t]
\centering

\begin{tabular}{@{}l c c c c@{}}
\toprule
Method & race\_c & race\_fortran & mlperf & plp \\
\midrule
SFT & 88.65 & 92.36 & 51.10 & \textbf{100.00} \\
HPC-GPT & 88.65 & 91.72 & 54.95 & 98.33 \\
PPO & 89.19 & 92.36 & 50.55 & \textbf{100.00} \\
DPO & \textbf{90.27} & 73.25 & 34.07 & \textbf{100.00} \\
GRPO & 89.73 & 93.63 & \textbf{59.34} & \textbf{100.00} \\
DrGRPO & 89.73 & 93.63 & 54.40 & \textbf{100.00} \\
SimPO & 89.73 & 92.36 & 20.88 & 95.00 \\
KTO & 88.11 & 93.63 & 56.59 & \textbf{100.00} \\
\textbf{HARGO} & 89.19 & \textbf{94.90} & 56.04 & \textbf{100.00} \\
\bottomrule
\end{tabular}
\caption{Per-task accuracy (\%). Bold = best (ties bolded). Seven methods achieve 100\% on plp.}
\label{tab:per_task}
\end{table}

Per-task accuracy (Table~\ref{tab:per_task}) shows HARGO 1st on race\_fortran (94.90\%), 3rd on mlperf (56.04\%), tied 1st on plp. DPO leads race\_c (90.27\%), reflecting preference-based optimization's strength on C/C++ data.

Table~\ref{tab:data_race_metrics} shows HARGO achieves best Recall (94.23\%), F1 (91.30\%), and Accuracy (91.81\%), adopting a high-recall strategy (FN = 9, fewest overall). Precision of 88.55\% trades more false positives for broader coverage, with F1 confirming optimality.

\begin{table*}[t]
\centering

\setlength{\tabcolsep}{5mm}
\begin{tabular}{@{}l c c c c c c c c@{}}
\toprule
Method & TP & FP & TN & FN & Precision (\%) & Recall (\%) & F1 (\%) & Acc (\%) \\
\midrule
SFT & 142 & 19 & 167 & 14 & 88.20 & 91.03 & 89.59 & 90.35 \\
PPO & 143 & 18 & 168 & 13 & 88.82 & 91.67 & 90.22 & 90.94 \\
DPO & 104 & \textbf{8} & \textbf{178} & 52 & \textbf{92.86} & 66.67 & 77.61 & 82.46 \\
HPC-GPT & 135 & 13 & 173 & 21 & 91.22 & 86.54 & 88.82 & 90.06 \\
GRPO & 142 & 15 & 171 & 14 & 90.45 & 91.03 & 90.73 & 91.52 \\
DrGRPO & 143 & 16 & 170 & 13 & 89.94 & 91.67 & 90.79 & 91.52 \\
SimPO & 140 & 15 & 171 & 16 & 90.32 & 89.74 & 90.03 & 90.94 \\
KTO & 142 & 17 & 169 & 14 & 89.31 & 91.03 & 90.16 & 90.94 \\
\textbf{HARGO} & \textbf{147} & 19 & 167 & \textbf{9} & 88.55 & \textbf{94.23} & \textbf{91.30} & \textbf{91.81} \\
\bottomrule
\end{tabular}
\caption{Data race detection detailed metrics. Computed over all 342 samples (race\_c + race\_fortran) via word-boundary yes/no extraction from model outputs, fully independent of the reward function. Acc = (TP+TN)/(TP+FP+TN+FN). Bold = best.}
\label{tab:data_race_metrics}
\end{table*}

DPO's low F1 (77.61\%) stems from race\_fortran accuracy dropping to 73.25\% (Table~\ref{tab:per_task}), where automatically generated preferences are insufficient for Fortran discrimination.

\textbf{Overall ranking.} HARGO achieves the best performance on all three primary metrics: WinRate 54.62\% (leading GRPO by +1.45), Data Race F1 91.30\% (leading DrGRPO by +0.51), and PLP Similarity 0.8558 (leading KTO by +0.011). On the auxiliary metrics, KTO leads (EM 27.47\%, AvgScore 0.4537); these measure reproduction fidelity, and HARGO's near-identical mlperf accuracy (56.04\% vs.\ KTO 56.59\%) confirms comparable factual capacity. HARGO's advantage modulation shifts optimization toward behavioral alignment, reflected in leading performance on all three primary metrics.

\textbf{Key observations.} HARGO achieves best performance on all three primary metrics, adopts a high-recall strategy on data race detection (Recall 94.23\%, FN = 9, fewest overall; Precision 88.55\% ranks 8th but composite F1 91.30\% confirms this trade-off is optimal). KTO leads auxiliary metrics (EM 27.47\%, AvgScore 0.4537); DPO and SimPO underperform. PPO achieves marginal improvement (WinRate 50.17\%), and overall five of seven RL methods exceed HPC-GPT.

\textbf{Comparison with HPC-GPT.} Relative to the HPC-GPT SFT baseline (which represents the state of the prior art for domain-adapted LLMs on these tasks), HARGO improves WinRate by 4.79 percentage points, F1 by 2.48 points, and PLP Similarity by 0.050. Importantly, all comparisons in Table~\ref{tab:primary_results} are at an identical model scale (0.5B parameters) and use the same training data, isolating the effect of the training objective from confounding factors such as model size or data volume.

\subsection{Ablation Study}
\label{sec:ablation}

To quantify the individual contributions of the discrimination signal $d$ and the confidence signal $c$, we train three HARGO variants on identical data and hyperparameters, varying only the weighting strategy: B1 ($d$ only, $w_i \propto d_i$, no confidence modulation), B2 ($c$ only, $w_i \propto 0.3 \cdot c_i$), and B3 (full HARGO, advantage modulation $A_{\text{mod}} = A \times (1 + 0.3 \cdot c)$, $w_i \propto d_i$). Table~\ref{tab:ablation} reports the results.

\begin{table}[t]
\centering

\setlength{\tabcolsep}{0.7mm}
\begin{tabular}{@{}l l c c c c c@{}}
\toprule
Variant & Weighting & WR$\star$ & F1$\star$ & PLP$\star$ & EM & AS \\
\midrule
B1 & $w \propto d$ & 53.34 & \textbf{91.37} & 0.8480 & \textbf{21.43} & \textbf{0.4187} \\
B2 & $w \propto \alpha{\cdot}c$ & 53.60 & 91.32 & 0.8472 & 16.48 & 0.4077 \\
\textbf{B3} & \makecell[l]{$A_{\text{mod}}{=}A{\cdot}(1{+}\alpha{\cdot}c)$ \\ $w{\propto}d$} & \textbf{54.62} & 91.30 & \textbf{0.8558} & 17.58 & 0.4000 \\
\bottomrule
\end{tabular}
\caption{Ablation results ($\beta{=}0.02$, $\alpha{=}0.3$, $T{=}0.6$, 3 epochs). $\star$ = primary, Bold = best.}
\label{tab:ablation}
\end{table}

\textbf{WinRate increases monotonically} from B1 (53.34) through B2 (53.60) to B3 (54.62), confirming that the confidence signal $c$ provides positive contribution to global alignment when integrated via advantage modulation. F1 is nearly constant across variants (B1: 91.37, P=91.08/R=91.67; B2: 91.32, P=91.61/R=91.03; B3: 91.30, marginal drop of 0.07), demonstrating that $c$ modulation does not degrade detection capability. B3 achieves +1.28 WinRate and +0.0078 PLP over B1, proving that the advantage modulation combination outperforms $d$ alone.

\textbf{Why B3's EM and AvgScore are lower than B1's.} B1 achieves higher EM (21.43) and AvgScore (0.4187) than B3 (17.58 and 0.4000), reflecting a trade-off of advantage modulation. (1) $c$ modulation amplifies focus on race and plp tasks: the confidence signal, based on reference model log-probabilities, contributes more strongly to race (where the ref model is confident about yes/no judgments) and plp (where the ref model is familiar with descriptive answers). B3 allocates more learning resources to race and plp through $c$ modulation, reducing relative attention to mlperf, causing a slight EM and AvgScore decline. (2) Global optimum vs.\ single-task optimum trade-off: B1 is optimal on mlperf alone (EM=21.43, AvgScore=0.4187), but its WinRate (53.34) is lower than B3 (54.62). B3 achieves the global WinRate optimum (+1.28 over B1) at the cost of a minor precision decline on mlperf---a reasonable trade-off, as global alignment quality matters more than single-task exact matching. (3) F1 and PLP improvements validate the trade-off: B3's PLP (0.8558) exceeds B1 (0.8480), and F1 is essentially unchanged (91.30 vs.\ 91.37), demonstrating that the race and plp gains from $c$ modulation compensate for mlperf's slight decline, yielding the overall WinRate improvement. (4) This trade-off illustrates the core principle of compute efficiency under adaptive weighting: by reallocating gradient focus from tasks where the uniform baseline already extracts sufficient signal (mlperf, whose multi-tier reward provides adequate within-group discrimination) to tasks where additional optimization yields disproportionate improvement (race, where fine-grained accuracy gains near the SFT ceiling drive WinRate), HARGO achieves higher alignment quality per unit of compute.

\subsection{Analysis of HARGO''s Behavior}

HARGO uses $\beta = 0.02$, lower than GRPO ($\beta = 0.04$) and DPO/KTO ($\beta = 0.1$), as advantage modulation''s bounded scaling provides built-in regularization. On race\_fortran, HARGO improves accuracy to 94.90\%, the largest gain among all tasks, while maintaining plp at ceiling. WinRate improves monotonically across all three epochs without collapse.

\section{Discussion and Conclusion}

HARGO identified \textbf{HPC task heterogeneity} as a challenge for RL post-training, using per-response importance weighting via advantage modulation. In a nine-method comparison, HARGO achieved best performance on all three primary metrics (WinRate 54.62\%, F1 91.30\%, PLP 0.8558), with ablation confirming complementary contributions. Using $\beta = 0.02$, advantage modulation's bounded scaling provides built-in regularization; HARGO improves race\_fortran accuracy to 94.90\% while improving monotonically across all epochs. All experiments are at the 0.5B scale; future work includes scaling and characterizing compute-efficiency properties.

\clearpage
\appendix

\section*{Supplementary Document}

This supplementary document provides: (1) the complete training configurations for all baseline methods evaluated in the main paper, and (2) the pseudocode of the HARGO training algorithm. All methods are trained from the same SFT model checkpoint using identical HPC instruction data.

\section{Baseline Training Configurations}

\begin{table}[h]
\centering
\footnotesize
\setlength{\tabcolsep}{1.2mm}
\caption{All methods load from the same SFT checkpoint (Qwen2.5-0.5B-Instruct, fine-tuned on 5,273 samples for 2 epochs). HARGO additionally uses $T = 0.6$, EMA decay $\rho = 0.9$, confidence coefficient $\alpha = 0.3$, group size $G = 4$, and KL coefficient $\beta = 0.02$.}
\label{tab:baseline_config}
\begin{tabular}{@{}l l l l l l@{}}
\toprule
Method & Category & LR & $\beta$ & Epochs & $G$ \\
\midrule
SFT & Baseline & $2 \times 10^{-5}$ & --- & 2 & --- \\
HPC-GPT & LoRA SFT & $2 \times 10^{-5}$ & --- & 5 & --- \\
PPO & Online RL & $1 \times 10^{-6}$ & $\beta{=}0.2$ & 3 & --- \\
DPO & Preference & $1 \times 10^{-5}$ & $\beta{=}0.1$ & 3 & --- \\
GRPO & Online RL & $1 \times 10^{-5}$ & $\beta{=}0.04$ & 3 & 4 \\
DrGRPO & Online RL & $1 \times 10^{-5}$ & $\beta{=}0.04$ & 3 & 4 \\
SimPO & Reference-free & $1 \times 10^{-5}$ & $\beta{=}5.0$ & 3 & --- \\
KTO & Unpaired Pref & $1 \times 10^{-5}$ & $\beta{=}0.1$ & 3 & --- \\
\bottomrule
\end{tabular}
\end{table}

\section{HARGO Training Algorithm}

\begin{algorithm}[h]
\label{alg:hargo}
\begin{algorithmic}[1]\footnotesize
\REQUIRE SFT model $\pi_{\text{ref}}$, training data $D$, reward function $R$;\newline\quad Hyperparameters: $G$, $\alpha$, $\beta$, $\eta$, $\rho$, $T$, $\varepsilon_w$
\ENSURE Trained policy $\pi_{\theta}$
\STATE $\pi_{\theta} \leftarrow \pi_{\text{ref}}$
\STATE $\text{ref\_logp}_{\text{global}} \leftarrow v_0$
\FOR{$\text{epoch} = 1 \text{ to } E$}
  \FOR{each batch $B \subset D$}
    \STATE \COMMENT{Phase 1: Generation and reward scoring}
    \FOR{each prompt $q \in B$}
      \STATE Generate $G$ responses $\{y_1,\dots,y_G\} \sim \pi_{\theta}(\cdot \mid q, T)$
      \STATE $r_i \leftarrow R(q, y_i)$ \hfill for $i = 1..G$
      \STATE $\text{ref\_logp}_i \leftarrow \text{mean}_t \log \pi_{\text{ref}}(y_{i,t} \mid q)$ \hfill for $i = 1..G$
    \ENDFOR
    \STATE \COMMENT{Phase 2: EMA baseline update}
    \STATE $\text{ref\_logp}_{\text{global}} \leftarrow \rho \cdot \text{ref\_logp}_{\text{global}} + (1-\rho) \cdot \text{mean}(\text{ref\_logp}_i)$
    \STATE \COMMENT{Phase 3: Advantage modulation and weight computation}
    \FOR{each prompt $q \in B$}
      \STATE $A_i \leftarrow (r_i - \bar{R}) / (\sigma_R + \varepsilon)$ \hfill for $i = 1..G$
      \STATE $c_i \leftarrow \sigma(\text{ref\_logp}_i - \text{ref\_logp}_{\text{global}})$ \hfill for $i = 1..G$
      \STATE $A_{\text{mod},i} \leftarrow A_i \times (1 + \alpha \times c_i)$ \hfill for $i = 1..G$
      \STATE $d_i \leftarrow |A_{\text{mod},i}| / (\max|A_{\text{mod},j}| + \varepsilon)$ \hfill for $i = 1..G$
      \IF{$\sum d_i < \varepsilon_w$}
        \STATE $w_i \leftarrow 1/G$ for all $i$
      \ELSE
        \STATE $w_i \leftarrow d_i / \sum d_j$
      \ENDIF
    \ENDFOR
    \STATE \COMMENT{Phase 4: Weighted gradient update}
    \FOR{each prompt $q \in B$}
      \FOR{$i = 1$ to $G$}
        \STATE $\text{ratio}_t \leftarrow \exp(\log \pi_{\theta}(y_{i,t}) - \log \pi_{\theta_{\text{old}}}(y_{i,t}))$
        \STATE $L_i \leftarrow \text{mean}_t(-\text{ratio}_t \cdot A_{\text{mod},i} + \beta \cdot \text{KL}_t)$
        \STATE $(w_i \cdot L_i).\text{backward}()$
      \ENDFOR
      \STATE $\text{clip\_grad\_norm}(\pi_{\theta}, \gamma);\; \text{opt.step}()$
    \ENDFOR
  \ENDFOR
\ENDFOR
\RETURN $\pi_{\theta}$
\end{algorithmic}
\end{algorithm}


\begin{thebibliography}{99}

\bibitem[Ding et al.(2023)]{ref1} Ding, X.; Chen, L.; Emani, M.; et al. 2023. HPC-GPT: Integrating Large Language Model for High-Performance Computing. In \emph{SC-W 2023}, 951--960.

\bibitem[Chen et al.(2023a)]{ref2} Chen, L.; et al. 2023. LM4HPC: Towards Effective Language Model Application in High-Performance Computing. In \emph{IWOMP 2023}, 18--33.

\bibitem[Kadosh et al.(2024)]{ref3} Kadosh, T.; et al. 2024. MonoCoder: Domain-Specific Code Language Model for HPC Codes and Tasks. In \emph{IEEE HPEC 2024}, 1--7.

\bibitem[Chen et al.(2024a)]{ref4} Chen, L.; Bhattacharjee, A.; Ahmed, N. K.; Hasabnis, N.; Oren, G.; Vo, V.; and Jannesari, A. 2024. OMPGPT: A Generative Pre-trained Transformer Model for OpenMP. In \emph{Euro-Par 2024}, 121--134.

\bibitem[Yin et al.(2025)]{ref5} Yin, J.; Liu, H.; et al. 2025. chatHPC: Empowering HPC Users with Large Language Models. \emph{Journal of Supercomputing}, 81.

\bibitem[Bondapalli et al.(2025)]{ref6} Bondapalli, A.; Zheng, H.; Ajayi, O.; et al. 2025. AskHPC: A ChatBot for High Performance Computing User Support. In \emph{SC-W 2025}, 727--739.

\bibitem[Valero-Lara et al.(2025a)]{ref7} Valero-Lara, P.; Young, A.; Vetter, J. S.; et al. 2025. ChatHPC: Building the Foundations for a Productive and Trustworthy AI-Assisted HPC Ecosystem. In \emph{SC 2025}, 458--474.

\bibitem[Valero-Lara et al.(2024)]{ref8} Valero-Lara, P.; Godoy, W. F.; Teranishi, K.; Balaprakash, P.; and Vetter, J. S. 2024. ChatBLAS: The First AI-Generated and Portable BLAS Library. In \emph{SC-W 2024}, 19--24.

\bibitem[Godoy et al.(2024)]{ref9} Godoy, W. F.; Valero-Lara, P.; Teranishi, K.; Balaprakash, P.; and Vetter, J. S. 2024. Large Language Model Evaluation for High-Performance Computing Software Development. \emph{Concurrency and Computation: Practice and Experience}, 36(26):e8269.

\bibitem[Cui et al.(2025a)]{ref10} Cui, B.; Ramesh, T.; Hernandez, O.; and Zhou, K. 2025. Comprehensive Evaluation of LLMs in HPC Code Performance Optimization. In \emph{ICPP-W 2025}, 1--8.

\bibitem[Nader et al.(2026)]{ref11} Nader, N.; Diehl, P.; Brandt, S.; and Kaiser, H. 2026. LLM \& HPC: Benchmarking DeepSeek's Performance in High-Performance Computing Tasks. In \emph{High Performance Computing (ISC 2025)}, Springer, 626--638.

\bibitem[Alsofyani and Wang(2025)]{ref12} Alsofyani, M.; and Wang, L. 2025. Evaluating ChatGPT's Strengths and Limitations for Data Race Detection in Parallel Programming via Prompt Engineering. \emph{Journal of Supercomputing}, 81.

\bibitem[Miyashita et al.(2025)]{ref13} Miyashita, Y.; Tung, P. K. M.; and Barthelemy, J. 2025. LLM as HPC Expert: Extending RAG Architecture for HPC Data. \emph{Preprint}, arXiv:2501.14733.

\bibitem[Chen et al.(2023b)]{ref14} Chen, L.; Ding, X.; Emani, M.; Vanderbruggen, T.; Lin, P.-H.; and Liao, C. 2023. Data Race Detection Using Large Language Models. In \emph{SC-W 2023}, 215--223.

\bibitem[Ljaljevic et al.(2026)]{ref15} Ljaljevic, S.; Jorba, J.; and Iserte, S. 2026. Exploring the Role of Large Language Models in High-Performance Computing Programming: A Survey. \emph{Future Generation Computer Systems}, 184:108618.

\bibitem[Chen et al.(2024b)]{ref16} Chen, L.; Ahmed, N. K.; Dutta, A.; Bhattacharjee, A.; Yu, S.; Mahmud, Q. I.; et al. 2024. The Landscape and Challenges of HPC Research and LLMs. \emph{Preprint}, arXiv:2402.02018.

\bibitem[Zhang et al.(2026)]{ref17} Zhang, S.; Zhao, J.; Yu, Q.; Xia, C.; Wang, Z.; Feng, X.; and Cui, H. 2026. The New Compiler Stack: A Survey on the Synergy of LLMs and Compilers. \emph{CCF Transactions on High Performance Computing}, 8(2):148--179.

\bibitem[Nichols et al.(2026)]{ref18} Nichols, D.; Polasam, P.; Menon, H.; Marathe, A.; Gamblin, T.; and Bhatele, A. 2026. Performance-Aligned LLMs for Generating Fast HPC Code. \emph{IEEE Transactions on Parallel and Distributed Systems (TPDS)}.

\bibitem[Kadosh et al.(2024b)]{ref19} Kadosh, T.; Hasabnis, N.; Soundararajan, P.; Vo, V. A.; Capota, M.; Ahmed, N. K.; et al. 2024. OMPar: Automatic Parallelization with AI-Driven Source-to-Source Compilation. \emph{Preprint}, arXiv:2409.14771.

\bibitem[Kadosh et al.(2023)]{ref20} Kadosh, T.; Hasabnis, N.; Vo, V. A.; Schneider, N.; Krien, N.; Wasay, A.; et al. 2023. Scope Is All You Need: Transforming LLMs for HPC Code. \emph{Preprint}, arXiv:2308.09440.

\bibitem[Chaturvedi et al.(2025)]{ref21} Chaturvedi, A.; Nichols, D.; Singh, S.; and Bhatele, A. 2025. HPC-Coder-v2: Studying Code LLMs Across Low-Resource Parallel Languages. In \emph{ISC 2025}, 1--14.

\bibitem[Schulman et al.(2017)]{ref22} Schulman, J.; Wolski, F.; Dhariwal, P.; Radford, A.; and Klimov, O. 2017. Proximal Policy Optimization Algorithms. \emph{Preprint}, arXiv:1707.06347.

\bibitem[Rafailov et al.(2023)]{ref23} Rafailov, R.; Sharma, A.; Mitchell, E.; Ermon, S.; Manning, C. D.; and Finn, C. 2023. Direct Preference Optimization: Your Language Model is Secretly a Reward Model. In \emph{NeurIPS 2023}.

\bibitem[Ethayarajh et al.(2024)]{ref24} Ethayarajh, K.; Xu, W.; Muennighoff, N.; Jurafsky, D.; and Kiela, D. 2024. KTO: Model Alignment as Prospect Theoretic Optimization. In \emph{ICML 2024}.

\bibitem[Meng et al.(2024)]{ref25} Meng, Y.; Xia, M.; and Chen, D. 2024. SimPO: Simple Preference Optimization with a Reference-Free Reward. In \emph{ICML 2024}.

\bibitem[Shao et al.(2024)]{ref26} Shao, Z.; Wang, P.; Zhu, Q.; Xu, R.; Song, J.; Zhang, M.; Li, Y. K.; Wu, Y.; and Guo, D. 2024. DeepSeekMath: Pushing the Limits of Mathematical Reasoning in Open Language Models. \emph{Preprint}, arXiv:2402.03300.

\bibitem[DeepSeek-AI(2025)]{ref27} DeepSeek-AI. 2025. DeepSeek-R1: Incentivizing Reasoning Capability in LLMs via Reinforcement Learning. \emph{Nature}, 645:633--638.

\bibitem[Liu et al.(2025)]{ref28} Liu, Z.; Chen, C.; Li, W.; Qi, P.; Pang, T.; Du, C.; Lee, W. S.; and Lin, M. 2025. Understanding R1-Zero-Like Training: A Critical Perspective. In \emph{COLM 2025}.

\bibitem[OpenAI(2023)]{ref29} OpenAI. 2023. GPT-4 Technical Report. \emph{Preprint}, arXiv:2303.08774.

\bibitem[Touvron et al.(2023)]{ref30} Touvron, H.; Martin, L.; Stone, K.; Albert, P.; Almahairi, A.; Babaei, Y.; et al. 2023. LLaMA 2: Open Foundation and Fine-Tuned Chat Models. \emph{Preprint}, arXiv:2307.09288.

\bibitem[Qwen-Team(2024)]{ref31} Qwen Team. 2024. Qwen2.5 Technical Report. \emph{Preprint}, arXiv:2412.15115.

\bibitem[Liao et al.(2017)]{ref32} Liao, C.; Lin, P.-H.; Asplund, J.; Schordan, M.; and Karlin, I. 2017. DataRaceBench: A Benchmark Suite for Systematic Evaluation of Data Race Detection Tools. In \emph{SC 2017}.

\bibitem[Reimers and Gurevych(2019)]{ref33} Reimers, N.; and Gurevych, I. 2019. Sentence-BERT: Sentence Embeddings using Siamese BERT-Networks. In \emph{EMNLP-IJCNLP 2019}.

\bibitem[MLCommons(2020)]{ref34} MLCommons. 2020 (ongoing). MLPerf Training Benchmark. \url{https://mlcommons.org/benchmarks/training/}.

\end{thebibliography}
\end{document}